\documentclass[runningheads]{llncs}
\usepackage[T1]{fontenc}
\usepackage{graphicx, verbatim}
\usepackage{hyperref}
\usepackage{color}

\usepackage{graphicx}
\usepackage{amsmath}
\usepackage{multirow}
\usepackage{float}
\usepackage{xspace}
\usepackage{overpic}
\usepackage{wrapfig}
\usepackage{amsfonts}
\usepackage{xcolor}
\usepackage{siunitx}
\usepackage{amssymb}
\usepackage{textcomp}
\usepackage{subcaption}

\newcommand{\err}[1]{{\fontsize{8pt}{8pt}\selectfont\textpm\thinspace#1}}
\begin{document}
    \title{TSegAgent: Zero-Shot Tooth Segmentation via Geometry-Aware Vision-Language
    Agents}
    \titlerunning{TSegAgent: Zero-Shot Tooth Segmentation}
    %
\author{Shaojie Zhuang\inst{1} \and
Lu Yin\inst{2}\thanks{Corresponding authors: Yuanfeng Zhou, Lu Yin.} \and
Guangshun Wei\inst{1} \and
Yunpeng Li\inst{3} \and
Xilu Wang\inst{2} \and
Yuanfeng Zhou\inst{1}*}
\authorrunning{S. Zhuang et al.}
%
\institute{Shandong University, Jinan 250010, China \and
University of Surrey, Guildford, Surrey, United Kingdom \and
King's College London, London, United Kingdom}



    \maketitle 

    \begin{abstract}
    Automatic tooth segmentation and identification from intra-oral scanned 3D models
    are fundamental problems in digital dentistry, yet most existing approaches
    rely on task-specific 3D neural networks trained with densely annotated
    datasets, resulting in high annotation cost and limited generalization to scans
    from unseen sources. Thus, we propose TSegAgent, which addresses these
    challenges by reformulating dental analysis as a zero-shot geometric
    reasoning problem rather than a purely data-driven recognition task. The key
    idea is to combine the representational capacity of general-purpose foundation
    models with explicit geometric inductive biases derived from dental anatomy.
    Instead of learning dental-specific features, the proposed framework leverages
    multi-view visual abstraction and geometry-grounded reasoning to infer tooth
    instances and identities without task-specific training. By explicitly
    encoding structural constraints such as dental arch organization and
    volumetric relationships, the method reduces uncertainty in ambiguous cases and
    mitigates overfitting to particular shape distributions. Experimental
    results demonstrate that this reasoning-oriented formulation enables
    accurate and reliable tooth segmentation and identification with low
    computational and annotation cost, while exhibiting strong generalization across
    diverse and previously unseen dental scans.
    \footnote{Code is available at: \hyperlink{https://github.com/znshje/TSegAgent}{https://github.com/znshje/TSegAgent}.}

\end{abstract}
    \section{Introduction}

Automatic tooth segmentation and identification from intra-oral scanned 3D models are essential components of digital dentistry, supporting a wide range of clinical applications such as orthodontic diagnosis, treatment planning, and prosthodontic design.
Most existing methods~\cite{lian2020deep,zhang2021tsgcnet,cui2021tsegnet,zhuang2023robust,lin2024dbganet,xi20253d} formulate these tasks as supervised learning problems and rely on task-specific 3D neural networks trained with densely annotated dental datasets like Teeth3DS~\cite{ben2022teeth3ds}.
While such approaches have shown promising performance under controlled settings, they require substantial annotation effort and often exhibit limited generalization when applied to scans acquired from unseen sources, scanners, or patient populations.

Vision backbones are advancing by improving the reception mechanism from local to global~\cite{ronneberger2015u,qi2017pointnet,qi2017pointnet++,wang2019dgcnn}. Transformer~\cite{vaswani2017attention,dosovitskiy2020image} and mamba~\cite{gu2024mambalineartimesequencemodeling,hatamizadeh2025mambavision} backbones change the way images are processed. Since large models like SAM~\cite{kirillov2023segany} became stronger in vision understanding, increasing studies adapted SAM to different areas and tasks.
Examples include SAM-Med3D \cite{wang2024sammed3d} and MedSAM~\cite{MedSAM}, that adapt general-purpose segmentation models to medical data, as well as conversational systems such as ChatIOS \cite{wu2025chatios} that explore vision-language interaction for intra-oral scan analysis. Recent IOSSAM \cite{Hua_IOSSAM_MICCAI2024} and 3DTeethSAM~\cite{lu20253dteethsamtamingsam23d} demonstrate great potential in SAM-based segmentation, but they require training a teeth detector or refiner, and lack the ability of segmentation error detection. With vision-language models advance, SAM3 \cite{carion2025sam3segmentconcepts} features a text-prompt segment anything model, enabling the training-free tooth segmentation on images.
These methods demonstrate the potential of foundation models to reduce annotation requirements and improve flexibility.
However, most existing approaches still treat tooth segmentation and identification as direct prediction problems, focusing on adapting large models to dental data rather than rethinking the underlying problem formulation.

In this work, we propose TSegAgent, a geometry-aware vision-language agent that reformulates tooth segmentation and identification as a zero-shot geometric reasoning problem.
Instead of learning dental-specific representations, TSegAgent combines general-purpose foundation models with explicit geometric inductive biases derived from dental anatomy.
The core idea is to leverage multi-view visual abstraction for instance extraction while enforcing anatomical consistency through geometry-grounded reasoning at the instance level.
By explicitly encoding dental arch structure, volumetric relationships, and symmetry priors, TSegAgent enables anatomically consistent tooth segmentation and identification without task-specific training.

This reasoning-oriented formulation offers two key advantages.
First, it significantly reduces annotation and training requirements while avoiding overfitting to particular model shapes or datasets, leading to strong generalization across dental scans from diverse and previously unseen sources.
Second, the integration of interpretable geometric priors reduces uncertainty in visually ambiguous cases, which is critical for clinical reliability.
Extensive experiments demonstrate that TSegAgent achieves accurate and consistent tooth segmentation and identification with low computational and annotation cost, highlighting the potential of geometry-grounded reasoning for practical dental AI systems. To sum up, our contributions include: (1) We propose a zero-shot framework for tooth instance segmentation and identification from intra-oral scanned 3D models, eliminating the need for task-specific training and dense annotations; (2) We design a geometry-aware and interpretable vision-language agent that integrates dental arch priors, multi-view visual evidence, and volumetric information to achieve accurate tooth classification; (3) We introduce explicit geometric reasoning and verification mechanisms that reduce ambiguity in tooth identification and enable robust generalization across dental models from arbitrary sources.

    \begin{figure}
    \centering
    \includegraphics[width=0.99\textwidth]{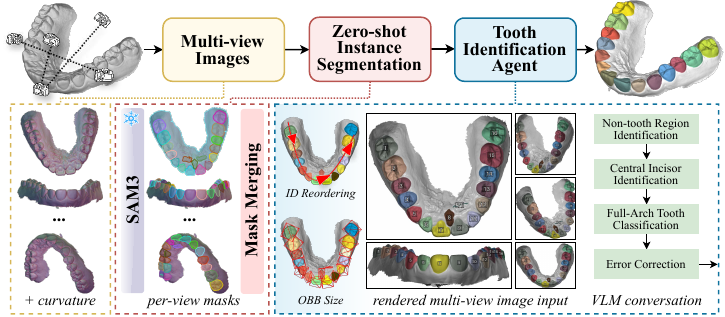}
    \caption{The pipeline of TSegAgent. Given an intra-oral scanned 3D model, we
    first perform multi-view rendering (with curvature) and apply SAM3 for zero-shot
    tooth instance segmentation. The resulting masks are merged into face-level instance
    labels, which are then reordered based on dental arch geometry. Finally, a
    vision-language agent identifies tooth instances by multi-round conversation
    and reasoning over visual cues and geometric constraints.}
    \label{fig:pipeline}
\end{figure}

\section{Method}

\subsection{Overview}

Given an intra-oral scanned (IOS) model, our goal is to automatically segment
individual tooth instances and assign FDI labels in a zero-shot manner. As illustrated
in Fig.~\ref{fig:pipeline}, TSegAgent consists of two components: multi-view
zero-shot tooth instance segmentation, and geometry-aware vision-language agent for
tooth identification. The entire pipeline avoids task-specific training and
global post optimization, relying instead on explicit geometric inductive biases
and reasoning to achieve interpretable results.

\subsection{Tooth Instance Segmentation by Multi-View Images}

The first component of TSegAgent focuses on extracting individual tooth
instances from an intra-oral scanned 3D model in a zero-shot manner. Since most general-purpose
segmentation models~\cite{kirillov2023segany,MedSAM} are designed for 2D images,
we convert the input 3D mesh into a set of multi-view renderings that capture
complementary perspectives of the dental arch. As shown in Fig.~\ref{fig:pipeline},
the rendering viewpoints are selected from five directions: perpendicular to the
occlusal plane, the patient's anterior-posterior direction, and the left and
right lateral sides. For each view, we additionally apply $10^{\circ}$ to
$30^{\circ}$ perturbations on the axes as new views to ensure the model surfaces
are covered as comprehensively as possible, while avoiding excessive inter-tooth
occlusion that would severely degrade the segmentation results.
Meanwhile, we add the curvature heatmap to the image to enhance fuzzy areas.
Each rendered image is then predicted independently using SAM3~\cite{carion2025sam3segmentconcepts}
with a text prompt ``tooth'', producing candidate instance masks without training.

These per-view predictions are projected back onto the 3D mesh. The
masks provide coarse instance proposals, and they are inherently unordered or incomplete
due to occlusion or view-dependent ambiguity. To obtain a unified instance
segmentation, we design a mask merging strategy that fuses multi-view predictions
into face-level instance labels, based on the overlap IoU metric between masks. Specifically,
for any two masks $M_{a}$ and $M_{b}$ from different views, if their IoU exceeds
the threshold, they are merged as one instance; If either $\lvert M_{a}\cap
M_{b}\rvert / \lvert M_{a}\rvert$ or $\lvert M_{a}\cap M_{b}\rvert / \lvert M_{b}
\rvert$ exceeds the threshold, we select the containment vs.\ non-containment relationship
that occurs most frequently across all views to determine whether $M_{a}$ contains
$M_{b}$ (or vice versa). We then use this consensus relationship to
differentiate instances.

\begin{figure}[t]
    \centering
    \begin{subfigure}
        {.25\linewidth}
        \centering
        \includegraphics[width=\linewidth]{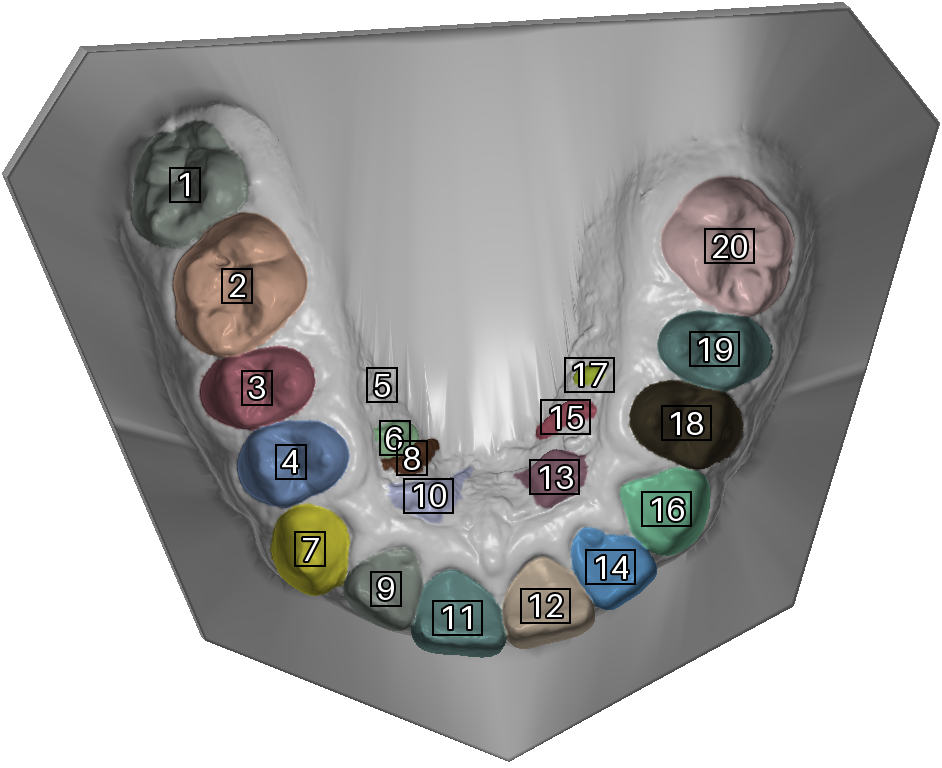}
        \caption{}
        \label{fig:challenge_cases:a}
    \end{subfigure}
    \hspace{2mm}
    \begin{subfigure}
        {.25\linewidth}
        \centering
        \includegraphics[width=\linewidth]{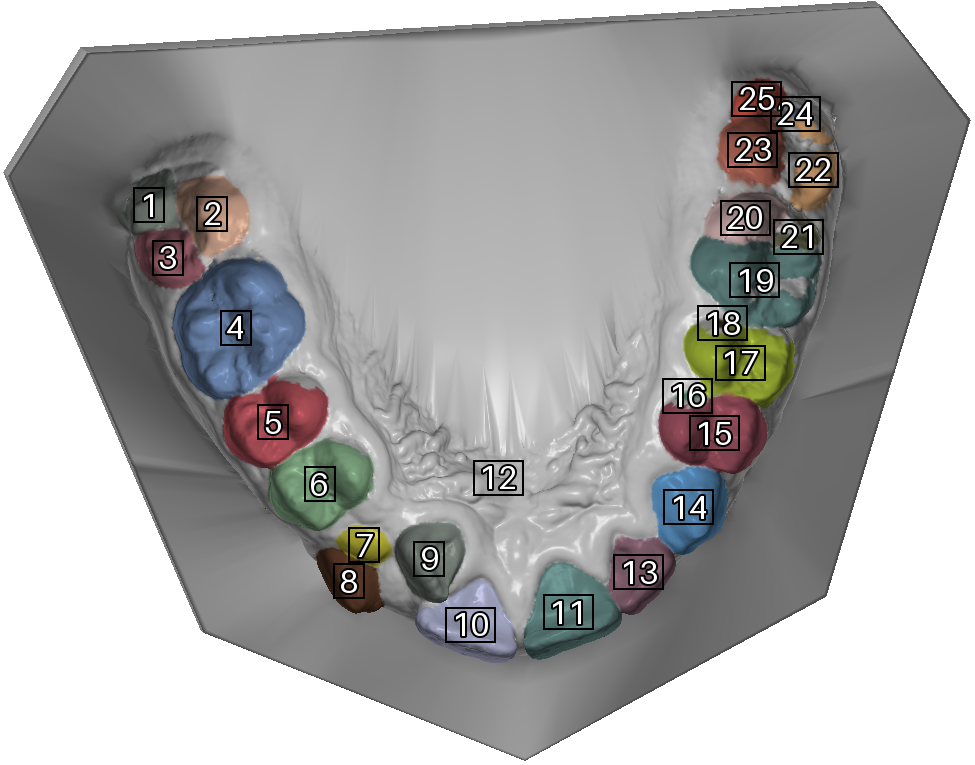}
        \caption{}
        \label{fig:challenge_cases:b}
    \end{subfigure}
    \hspace{2mm}
    \begin{subfigure}
        {.25\linewidth}
        \centering
        \includegraphics[width=\linewidth]{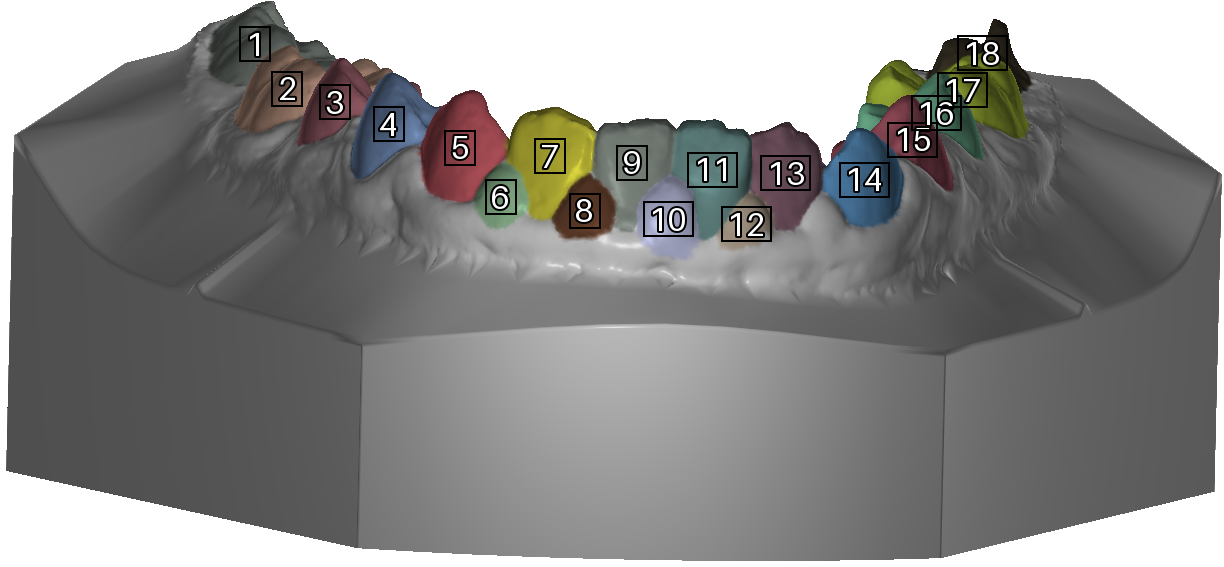}
        \caption{}
        \label{fig:challenge_cases:c}
    \end{subfigure}
    \caption{Typical challenging cases for tooth classification, including (a) non-tooth
    regions, (b) tooth over-segmentation into multiple instances due to
    occlusion or complex morphology, and (c) small gingival papilla between adjacent
    teeth that may be misclassified as tooth instances.}
    \label{fig:challenge_cases}
\end{figure}

\subsection{Tooth Identification Agent}

Despite the mask merging step, challenging cases still remain, including non-tooth
regions (e.g., gingiva), over-segmentation of teeth, and small gingival papillae
between adjacent teeth (Fig.~\ref{fig:challenge_cases}). By carefully designing
the text prompts and the image inputs, we leverage the capabilities of the vision-language
models (VLM) to eliminate these error cases.

\subsubsection{Instance ID Reordering.}

The tooth instances obtained from SAM3 are inherently unordered, which poses a challenge
for anatomically consistent tooth identification. To provide a structural
reference for reasoning, we perform instance ID reordering based on explicit dental
arch priors. For each segmented tooth instance, we compute its geometric centroid
from the associated mesh faces. All centroids are projected onto the XY-plane, where
a second-order polynomial curve is fitted to approximate the dental arch. Each
instance is assigned a scalar parameter corresponding to its closest point on
the fitted curve. By sorting instances according to this parameter, we reorder
instance IDs into a contiguous sequence that reflects their spatial arrangement
along the dental arch.

This reordering step encodes relative tooth positions in a geometry-driven
manner and enables subsequent reasoning processes to exploit sequential and
symmetry constraints without relying on learned positional embeddings.

\subsubsection{Multi-round Conversation-based Tooth Identification Agent.}

After instance reordering, tooth identification is performed by a geometry-aware
vision-language agent through a multi-round conversation. As illustrated in Fig.~\ref{fig:pipeline},
the dental mesh is rendered from selected views as image inputs, with temp tooth
ID printed on tooth instances. Rather than predicting all tooth labels in a single
pass, the agent decomposes the identification task into a sequence of
interpretable sub-tasks, each guided by explicit geometric and anatomical constraints.
Meanwhile, we add a 3-dimensional size of each tooth oriented bounding-box (OBB)
into text prompts, for VLM to easily evaluate tooth shapes. This design allows the
agent to progressively reduce ambiguity and improve robustness in zero-shot settings.
Detailed text prompt design can be found in our source codes.

\paragraph{Non-tooth Region Identification.}
In the first reasoning round, the agent identifies non-tooth instances, such as gingiva
or interdental papilla. The prompt is designed to check the size, position, and
shape of each tooth to identify non-tooth instances. Most challenging cases shown
in Fig.~\ref{fig:challenge_cases} can be effectively filtered out at this stage,
with the help of provided geometric clues. Non-tooth instances are excluded from
subsequent identification steps, simplifying the reasoning space and reducing potential
error propagation.

\paragraph{Central Incisor Identification.}
In the second reasoning round, the agent identifies central incisors, which
serve as a critical anatomical reference due to their bilateral symmetry and
central position along the dental arch. In the prompt, we guide the VLM to
determine the IDs of the central incisors based on dental arch symmetry and size
correspondence, rather than relying on left–right pixel balance in the image.


\paragraph{Full-Arch Tooth Classification.}
Conditioned on the reordered instance sequence and the identified central incisors,
VLM infers tooth identities by reasoning over relative position, morphological characteristics,
and volumetric relationships. We add detailed tooth features, common errors, and
output formats in text prompt. This sequential formulation enables the agent to
exploit dental arch continuity and avoid implausible label assignments.

\paragraph{Error Detection and Correction.}
During and after full-arch classification, the agent performs explicit error
detection based on bilateral symmetry, sequential consistency along the dental arch,
and relative volumetric similarity between corresponding teeth. When violations of
these constraints are detected, the agent is prompted to correct its previous
predictions through additional conversational rounds. This self-correction enables recovery from early errors without external supervision or post-hoc optimization, with segmentation gains mainly from non-tooth filtering and over-segmentation removal and error correction chiefly improving FDI accuracy.

After the multi-round conversation, the agent outputs the tooth ID to FDI label
mapping for each segmented instance, completing the zero-shot tooth identification
process.

    \section{Experiments}

\begin{figure}[t]
    \centering
    \includegraphics[width=0.99\textwidth]{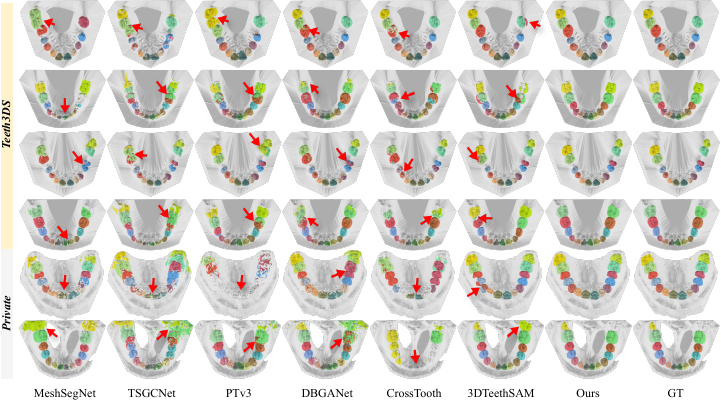}
    \caption{Qualitative results of candidate methods, from Teeth3DS and private
    dataset.}
    \label{fig:comparison}
\end{figure}

\subsection{Dataset and Settings}

We evaluate candidate methods on the Teeth3DS dataset~\cite{ben2022teeth3ds},
which contains 1200 3D tooth models. Each model is annotated with an FDI label for
each vertex. Meanwhile, to evalutate the generalizability of the methods, we
also test them on a private dataset collected from our collaborating dental
clinic, which consists of 340 3D tooth models in the testing set with the same annotation
scheme as Teeth3DS. The private dataset is more challenging than Teeth3DS, as most
are scanned plaster models with more severe defects, interference, and artifacts.
For each candidate methods, we train them on the training set of Teeth3DS from scratch
until convergence. Then we evaluate their performance on the testing set of
Teeth3DS and the private dataset.
TSegAgent is compatible with various VLM services, and use ``ChatGPT 5.2'' in our
experiments.

\def\oursPerformance{96.23\err{7.7} & 96.40\err{1.6} & 97.16\err{8.5} & 86.67}
\begin{table}[t]
    \centering
    \caption{Performance comparison on Teeth3DS and private dataset. The best results
    are highlighted in bold, while the second best results are underlined.}
    \label{tab:comparison}
    \begin{tabular}{|c|c|c|c|c|c|}
        \hline
        Method                                         & mIoU                        & TLA                         & TSA                        & TIR                         & TIR$_{=1}$        \\
        \hline
        \multicolumn{6}{|l|}{\textit{Teeth3DS Dataset}} \\
        \hline
        MeshSegNet~\cite{lian2020deep}                 & 76.46\err{13.2}             & 90.98\err{11.8}             & 89.70\err{4.5}             & 95.09\err{14.3}             & 83.33             \\
        PTv3~\cite{wu2024ptv3}                         & 83.61\err{11.6}             & \underline{94.45\err{8.3}}  & 95.22\err{1.5}             & 95.29\err{12.1}             & 79.67             \\
        TSGCNet~\cite{zhang2021tsgcnet}                & 53.45\err{14.4}             & 82.80\err{10.8}             & 85.23\err{3.7}             & 85.99\err{20.3}             & 51.17             \\
        DBGANet~\cite{lin2024dbganet}                  & 83.51\err{13.1}             & 95.37\err{7.0}              & 96.09\err{1.4}             & 95.80\err{13.2}             & \underline{86.50} \\
        CrossTooth~\cite{xi20253d}                     & 47.51\err{29.0}             & 63.57\err{47.2}             & 72.29\err{34.69}           & 70.21\err{40.0}             & 49.17             \\
        3DTeethSAM~\cite{lu20253dteethsamtamingsam23d} & \underline{92.03\err{9.5}}  & 93.64\err{13.9}             & \textbf{97.12\err{2.5}}    & 96.03\err{12.0}             & 85.00             \\
        TSegAgent (Ours)                               & \textbf{93.37\err{2.8}}     & \textbf{96.40\err{7.4}}     & \underline{96.76\err{1.5}} & \textbf{97.46\err{7.6}}     & \textbf{87.17}    \\
        \hline
        \multicolumn{6}{|l|}{\textit{Private Dataset}}  \\
        \hline
        MeshSegNet~\cite{lian2020deep}                 & 38.69\err{22.7}             & 80.83\err{36.8}             & 85.69\err{9.1}             & 76.33\err{24.4}             & \underline{39.41} \\
        PTv3~\cite{wu2024ptv3}                         & 36.42\err{19.3}             & 81.40\err{34.8}             & 86.67\err{4.6}             & 60.84\err{32.7}             & 21.76             \\
        TSGCNet~\cite{zhang2021tsgcnet}                & 28.75\err{10.8}             & 68.21\err{12.8}             & 80.70\err{4.9}             & 53.63\err{22.7}             & 3.83              \\
        DBGANet~\cite{lin2024dbganet}                  & 41.59\err{17.4}             & \underline{96.60\err{10.7}} & 91.10\err{4.2}             & 61.83\err{26.9}             & 22.06             \\
        CrossTooth~\cite{xi20253d}                     & 35.17\err{22.1}             & 34.48\err{46.8}             & 87.68\err{6.6}             & 46.99\err{29.1}             & 6.76              \\
        3DTeethSAM~\cite{lu20253dteethsamtamingsam23d} & \underline{81.42\err{14.1}} & 70.98\err{26.3}             & \underline{94.63\err{8.5}} & \underline{82.42\err{19.6}} & 37.06             \\
        TSegAgent (Ours)                               & \textbf{82.10\err{11.1}}    & \textbf{96.68\err{1.8}}     & \textbf{95.99\err{3.1}}    & \textbf{85.41\err{17.9}}    & \textbf{51.96}    \\
        \hline
    \end{tabular}
\end{table}

\subsection{Metrics}

We use the same metrics as in Teeth3DS benchmarks~\cite{ben2022teeth3ds}: TLA, TSA
and TIR. Additionally, we report $\mathrm{TIR}_{=1}$, the proportion of samples
where all instances are correctly identified. For each GT tooth $P_{i}$ with centroid
$\mathbf{c}_{i}$ and size $\mathbf{s}_{i}$, we match it to the closest predicted
tooth $\hat{P}_{\pi(i)}$. The metrics are:
\begin{equation}
    \begin{aligned}
        \mathrm{TLA} & = \exp(-\frac{1}{N}\sum_{i=1}^{N}\left\|\frac{\mathbf{c}_i-\hat{\mathbf{c}}_{\pi(i)}}{\mathbf{s}_i}\right\|_{2}), \mathrm{TSA}= \mathrm{F1}\!\left([\mathbf{y}^{gt}\neq 0],\,[\mathbf{y}^{pred}\neq 0]\right), \\
        \mathrm{TIR} & = \frac{1}{N}\sum_{i=1}^{N}\!\left( \ell_{i}=\hat{\ell}_{\pi(i)}\right), \mathrm{TIR}_{=1}= \frac{1}{M}\sum_{m=1}^{M}\!\left(\mathrm{TIR}^{(m)}=1\right).
    \end{aligned}
\end{equation}

Additionally, we report the mean Intersection over Union (mIoU) for tooth-wise
segmentation performance. For each GT tooth, the IoU is computed with the
predicted tooth with the same label.

\subsection{Comparison}

Results are shown in Fig.~\ref{fig:comparison} and Tab.~\ref{tab:comparison}. On
the Teeth3DS dataset, TSegAgent achieves competitive performance and outstanding
stability across the metrics. Compared to training-based methods, TSegAgent demonstrates
better ability of instance differentiation. Especially, compared
to 3DTeethSAM which is also based on SAM, TSegAgent preserves better instance
integrity while achieving higher classification accuracy.

More importantly, on the private dataset with a substantial domain
shift, TSegAgent demonstrates significantly stronger generalization capability.
Compared to supervised methods such as 3DTeethSAM and DBGANet, which exhibit notable
performance degradation under cross-domain conditions, TSegAgent maintains
robust overall accuracy, highlighting the reliability of
the proposed reasoning framework in challenging real-world scenarios.

\subsection{Ablation Studies}

We conduct ablation experiments on the Teeth3DS dataset to analyze the contribution
of each component in TSegAgent in Tab.~\ref{tab:ablation_settings}. The segmentation-only
baseline (\textit{w/o} VLM) evaluates initial segmentation performance. Although
it achieves a high TSA score, TSA does not account for instance-level
distinguish; therefore, incorrect segmentations may still lead to an improvement
in TSA but low TLA. Introducing VLM significantly improves segmentation quality,
as VLM will wipe out any mis- or over-segmentation. Enabling ID reordering (Arch)
further improves TIR by providing a consistent positional reference along the
dental arch. Adding volumetric cues (OBB) enhances structural awareness and improves
robustness in distinguishing anatomically similar or bad teeth. Finally, decomposing
tooth identification into structured reasoning steps effectively reduces ambiguity
and enforces anatomical consistency, increasing TIR to 97.46\% and TIR$_{=1}$ to
87.17\%. Overall, the full TSegAgent configuration achieves the best performance,
validating the complementary benefits of geometric inductive biases and multi-round
reasoning.

\begin{table}[t]
    \centering
    \caption{Ablation study of segmentation and reasoning components in
    TSegAgent, with performance evaluated on Teeth3DS dataset.}
    \label{tab:ablation_settings}
    \begin{tabular}{|cccc|c|c|c|c|c|}
        \hline
        \multicolumn{4}{|c|}{Settings} & \multicolumn{5}{c|}{Metrics} \\
        \hline
        VLM                            & Conv.                       & Arch       & OBB        & mIoU                        & TLA                        & TSA                        & TIR                     & TIR$_{=1}$     \\
        \hline
        -                              & -                           & -          & -          & -                           & 75.21\err{27.1}            & \underline{96.45\err{1.7}} & -                       & -              \\
        \checkmark                     & \checkmark                  & -          & -          & 85.49\err{14.2}             & 91.67\err{15.2}            & 95.59\err{4.4}             & 89.64\err{18.3}         & 66.72          \\
        \checkmark                     & \checkmark                  & \checkmark & -          & \underline{86.20\err{16.4}} & 92.26\err{13.6}            & 96.02\err{2.1}             & 90.43\err{20.6}         & 74.83          \\
        \checkmark                     & -                           & \checkmark & \checkmark & 83.14\err{21.3}             & \underline{95.73\err{8.9}} & 96.27\err{1.7}             & 86.36\err{26.0}         & 71.17          \\
        \checkmark                     & \checkmark                  & \checkmark & \checkmark & \textbf{93.37\err{2.8}}     & \textbf{96.40\err{7.4}}    & \textbf{96.76\err{1.5}}    & \textbf{97.46\err{7.6}} & \textbf{87.17} \\
        \hline
    \end{tabular}
\end{table}

    \section{Conclusion}

In this work, we presented TSegAgent, a geometry-aware vision-language framework
for zero-shot tooth segmentation and identification, which highlights the
potential of integrating foundation models with structured geometric reasoning
for practical dental AI applications. Future work may explore extending the proposed
reasoning framework to other anatomical structures and incorporating additional domain-specific
priors to further enhance robustness and clinical applicability.


\subsubsection{Disclosure of Interests.} The authors have no competing interests to declare that are relevant to the content of this article.

    %
    %
    \bibliographystyle{splncs04}
    \bibliography{Paper-0737}
\end{document}